\documentclass[10pt,twocolumn,letterpaper]{article}

\usepackage{cvpr}              
\usepackage{graphicx}
\usepackage{amsmath}
\usepackage{amssymb}
\usepackage{booktabs}
\usepackage[pagebackref, breaklinks, colorlinks]{hyperref}
\hypersetup{
     breaklinks=true,
     colorlinks   = true,
     linkcolor    = blue,
     urlcolor     = blue
}
\usepackage[capitalize]{cleveref}

\usepackage{xcolor}
\usepackage{bbding}
\usepackage{pifont}
\usepackage{array}
\usepackage{multirow}
\usepackage{multicol}
\usepackage{tikz}

\usepackage{nicematrix}
\usepackage[percent]{overpic}

\author{
Rahul Goel$^{1}$\\
\and
Dhawal Sirikonda$^{1}$\\
\and
Rajvi Shah$^{2}$\\
\and
PJ Narayanan$^{1}$\\
\and
$^{1}$CVIT, KCIS, IIIT Hyderabad\\
\and
$^{2}$Meta Reality Labs, USA\\
}
\def\cvprPaperID{6}
\def\confName{CVPR}
\def\confYear{2023}
\begin{document}
{
    \title{FusedRF: Fusing Multiple Radiance Fields for Fast Rendering with Lower Memory Requirements.}
    \title{FusedRF: Fusing Multiple Radiance Fields}
    
    \twocolumn[{
    \renewcommand\twocolumn[1][]{#1}
    \maketitle
    \begin{center}
    \includegraphics[width=0.48\linewidth]{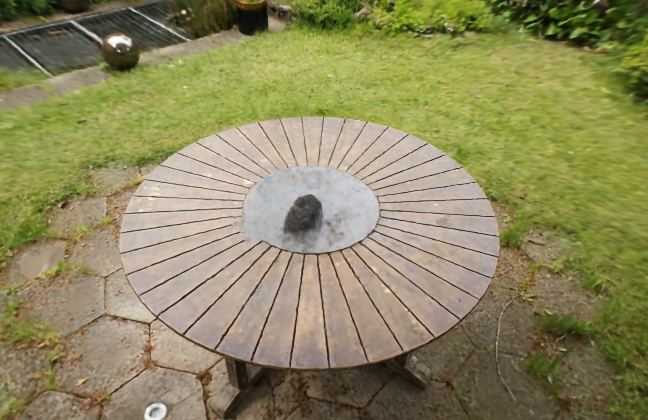}
    \includegraphics[width=0.48\linewidth]{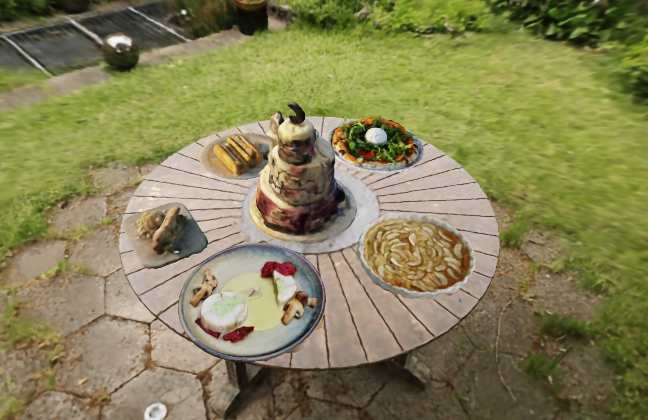}
    \captionof{figure}{\textit{NeRF-Feast:} Our method can be used to create a single composed representation. Here, we show a feast prepared on the table from the garden scene from MipNeRF360 \cite{mipnerf360}. Since we have a single fused representation, the scene is rendered at the cost of a single RF representation while also occupying a memory footprint of a single RF. Scene components are picked from \cite{Sketchfab}}
    \label{fig:teaser}
\end{center}
    }]
    
    \maketitle
    \crefname{section}{Sec.}{Secs.}
\crefname{table}{Tab.}{Tabs.}
\crefname{figure}{Fig.}{Figs.}
\crefname{equations}{Eq.}{Eqs.}

\newcommand{\crossmark}{{\color{red}\ding{56}}}
\newcommand{\tickmark}{{\color{green}\ding{52}}}

\newcolumntype{N}{@{}m{0pt}@{}}

\newcommand\ExtraSep
{\dimexpr\cmidrulewidth+\aboverulesep+\belowrulesep\relax}

\newcommand{\sqboxs}{1.2ex}
\newcommand{\sqboxf}{0.6pt}
\newcommand{\sqbox}[1]{\textcolor{#1}{\rule{\sqboxs}{\sqboxs}}}
\newcommand{\sqboxEmpty}[1]{%
  \begingroup
  \setlength{\fboxrule}{\sqboxf}%
  \setlength{\fboxsep}{-\fboxrule}%
  \textcolor{#1}{\fbox{\rule{0pt}{\sqboxs}\rule{\sqboxs}{0pt}}}%
  \endgroup
}
\newcommand{\tikzcircle}[2][red,fill=red]{\tikz[baseline=-0.5ex]\draw[#1,radius=#2] (0,0) circle ;}%

\definecolor{red}{HTML}{E51400}
\definecolor{light_red}{HTML}{F8CECC}
\definecolor{blue}{HTML}{0050EF}
\definecolor{light_blue}{HTML}{DAE8FC}
    
    \begin{abstract}
{
    Radiance Fields (RFs) have shown great potential to represent scenes from casually captured discrete views. Compositing parts or whole of multiple captured scenes could greatly interest several XR applications. Prior works can generate new views of such scenes by tracing each scene in parallel. This increases the render times and memory requirements with the number of components. In this work, we provide a method to create a single, compact, {\em fused} RF representation for a scene composited using multiple RFs. The fused RF has the same render times and memory utilizations as a single RF. Our method distills information from multiple teacher RFs into a single student RF while also facilitating further manipulations like addition and deletion into the fused representation.
}
\end{abstract}

    \section{Introduction}
{
    \label{sec:intro}
    The use of Radiance Fields (RF) to represent scenes has been of great interest lately. Given a discrete set of posed images of the scene, NeRF\cite{nerf} utilizes simple MLPs to encode the scene allowing the synthesis of novel views. While MLPs encoded the scene faithfully, the training time to learn the representation was high, in the range of 8-10 hours. Subsequent efforts like Plenoxels, DVGO \cite{plenoctrees, dvgo, dvgov2} exploited explicit 3D lattice structure to represent the scene and reduced the scene learning times to $\sim$10 minutes. Later extensions like TensoRF \cite{tensorf} compressed the volumetric lattice representation while maintaining low learning times.

    As the efficiency of scene representation improved, efforts were made towards editing the geometric content of the RFs. While efforts like \cite{nerf_inpaint_1, nerf_inpaint_2} aimed at removing the existing content and inpaint the missing regions, works like \cite{cagenerf, neumesh,nerfshop, nerfshopplusplus, neuphysics, spidr, edittoponerf,partnerf, neuman, confies} aimed at deforming the RF using cage-based and topological deformations.
    
    Unlike these, we are interested in compositional editing, in which objects and parts from multiple RFs are combined into a single, compact, {\em fused} RF representation. Compositing for view generation via parallel tracing rays into multiple RF scenes has been attempted before \cite{d2nerf, compositional1, compositional2, compositional3, compositional4, compositional5, compositional6, control_nerf}. As multiple component RFs that make up the scene need to be kept simultaneously, the memory footprint and rendering time increase in this method. Most prior works use the compact neural representation of RFs as a result. On the other hand, Control-NeRF \cite{control_nerf} uses full volumetric lattice representation with a common rendering network for all component RFs. This approach demands higher memory, specifically when more than two scenes are to be composited. Additionally, the render times are proportional to the number of RFs that make up the scene due to the need to shoot a large number of rays into each RF.

    High memory requirements and rendering times are undesirable for use cases of using the captured RF content in virtual environments, creating extended (XR) and augmented reality (AR) applications. This use requires lower memory and fast rendering of the final composited scene. One way to achieve this goal is to create a single representation by training an RF from scratch using a large number of rendered RGB views. The resulting radiance field will be a single, compact, composited RF with similar memory footprints and render times as a single RF. This approach is, however, inefficient as it involves complete view generation and retraining which could be expensive and slow. Given that all parts in the composited scene already use RF representation, is there a way to {\em fuse} them directly into an RF representation without expensive RGB retraining?

    In this paper, we propose a method to {\em fuse} multiple RFs into a single one using distillation on their RF  \cite{pvd}. Our distillation is similar to the step used by PVD to translate one type of RF to another \cite{pvd}. We iteratively fuse the source RFs (with affine composition) into a single, fused  RF. Distillation can be performed directly on the $\sigma$ and color values stored in the RF quickly in a supervised manner for acceptable results. The quality can be improved using a few additional iterations of RGB-based training at the end. The fused RF representation has the same memory requirements and rendering speed as a single RF. \cref{tab:works} presents the features of our system relative to others that can also do view generation of composited scenes using parallel ray tracing.

    \begin{table}
{
    \begin{NiceTabular}{lcccc}[hvlines, corners=NW]
    \Block{4-1}{Works} & \Block{1-2}{Native} & & \Block{1-2}{Composition} \\
    & \Block{3-1}{\footnotesize Small\\ \footnotesize Memory\\ \footnotesize Footprint} & \Block{3-1}{ \footnotesize Short\\ \footnotesize Training\\ \footnotesize Time} & \Block{3-1}{ \footnotesize Small\\ \footnotesize Memory\\ \footnotesize Footprint} & \Block{3-1}{\footnotesize Short\\ \footnotesize 
 Render\\ \footnotesize  Times} \\
    \\
    \\
    NeRF\cite{nerf}                 & \tickmark  & \crossmark & \tickmark  & \crossmark\\
    D$^2$NeRF\cite{d2nerf}          & \tickmark  & \crossmark & \tickmark  & \crossmark\\
    PlenOxels\cite{plenoxels}    & \crossmark & \tickmark  & \crossmark & \crossmark\\
    DVGO\cite{dvgo}                 & \crossmark & \tickmark  & \crossmark & \crossmark\\
    InstantNGP\cite{instantngp}    & \crossmark & \tickmark\tickmark  & \crossmark & \tickmark\tickmark\\
    CNeRF\cite{control_nerf}  & \crossmark & \tickmark  & \crossmark & \crossmark\\
    TensoRF\cite{tensorf}           & \tickmark  & \tickmark  & \tickmark  & \crossmark\\
    PVD\cite{pvd}                   & \tickmark  & \tickmark  & \crossmark  & \crossmark\\
    FusedRF (Ours)                  & \tickmark  & \tickmark  & \tickmark  & \tickmark\\
    \end{NiceTabular}
    \caption{\label{tab:works} While works like NeRF and D$^2$NeRF struggle in render times, works like DVGO, Plenoxel, and ControlNeRF additionally also demand high memory. Leading to the infeasibility of the composition of more than a few scenes. Our method efficiently fuses the RFs and maintains memory, and renders times to a single RF}\vspace{-5mm}
}
\end{table}
    
}

    \begin{figure*}[h]
\begin{center}
    \includegraphics[width=\linewidth]{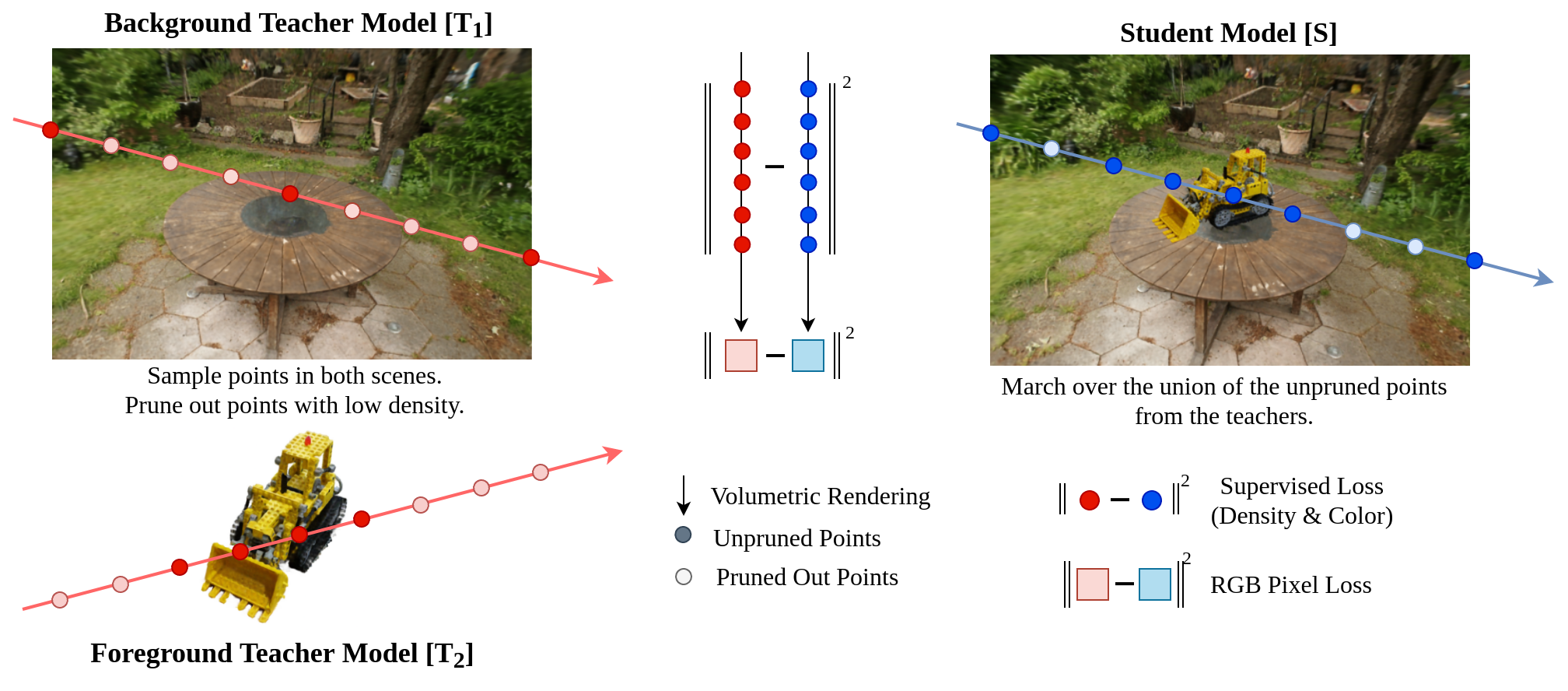}
    \captionof{figure}{\label{fig:sysdiag}\textit{FusedRF:} Method to fuse multiple RFs into single presentation. We shot the same ray into both teachers $T_1$, $T_2$ (left) created using ISRF \cite{isrf} to distill a combined Student $S$(right). For every sampled point on a ray of both RFs we prune out ones with lower densities and apply supervised distillation losses to obtain faster convergence. Followed by a few iterations of Pixel Loss on alpha-composited RGB for smoothening. By fusing multiple RFs into a single representation, we reduce rendering and memory overheads of composition. Both scenes \textsc{Garden} and \textsc{Lego} are picked from the real-world \cite{mipnerf360} dataset.}
    \vspace{-8mm}
\end{center}
\end{figure*}

    \section{Related Work}
{
    \label{sec:rel_work}
    Due to space constraints, we restrict this section to the following relevant works.
    
    \noindent\textbf{Editing Radiance Fields:} Several works have been proposed to edit Radiance Fields. The works like \cite{neural_pil, nerfactor, extracting_triangle, nerd, nerv2020} aim to edit the appearance by manipulating the light transport equation, works like \cite{stylizednerf_cvpr, style_implicit_wacv, snerf_siggraph} utilize image-based priors for the stylization of radiance fields. On the other hand works like \cite{neumesh, cagenerf, nerfshop, nerfshopplusplus} work on the deformation of radiance fields, and few works concentrate on compositing and creating a novel scene from multiple individually captured scenes\cite{d2nerf, compositional1, compositional2, compositional5, control_nerf}. We aim to create novel scenes by compositing captured scenes that have applications in XR.

    \noindent\textbf{Segmentation of Scene Content:} Often composition of these scenes involves partial addition of semantically correct content captured in one scene to another. Obtaining such semantically correct partial scene content is tackled in recent works either by fusing semantic features\cite{isrf, DFF, N3F} into RFs or propagating multi-view 2D masks or labels into the radiance field representations\cite{semanticnerf, spinnerf}. For this task, we choose \cite{isrf}, which is known to present the most accurate segmentation (extraction) of radiance fields.

    \noindent\textbf{Fusion of multiple Radiance Fields:} A simple composition using affine transformation has been attempted by works like D$^2$NeRF\cite{d2nerf} where compact Neural-RFs were employed to obtain the desired composition. The desired composition increases memory footprints and renders times in proportion to the number of scenes involved. In the case of explicit volumetric lattice, though the partial content retrieval is more accurate, the memory requirement grows rapidly, leading to infeasibility when more than a few scenes are being composited. Hence, a fused representation of composition is desired owing to memory footprints and render times as that of a single RF.

    \noindent\textbf{Efficient Fusion:}Works like PVD\cite{pvd} provide methodologies for faster distillation of one RF to another in a supervised setting. But the work is limited to a single RF representation. We draw ideas from this work and build a FusedRF representation that is compact and easy to render.
}
    \vspace{-3mm}
\section{Method}
{
    \label{sec:method}
    \cref{fig:sysdiag} shows an overview of our FusedRF method. The following through \cref{sub_sec:comp_nerf,sub_sec:fusion,sub_sec:fast_conv} will detail the compositional, fusion, and convergence aspects of the proposed methodology, respectively.
    
    \subsection{Composition of Radiance Fields}
    {
        \label{sub_sec:comp_nerf}
        The composition of two distinct radiance fields can be performed by altering the compositional aspect of the volumetric rendering equation \cite{nerf, dvgo}. For simplicity, let's assume we have to composite only two radiance fields ($RF^1$ and $RF^2$). Every ray is shot and sampled similarly in both RFs. For every sampled point along the ray, the activated volumetric density $(\alpha: \alpha^{I} = 1 - e^{-\sigma^{I} \delta^{I}})$ is calculated, where $I$ corresponds to the respective Radiance Field ($RF^{I}$)\cite{dvgo}. The point with a higher $\alpha$ value is chosen for a contribution towards the rendering of color and density. The resultant RF is considered ground truth for scene composition.
        It can be observed that the tracing and sampling of the same ray twice are redundant; subsequently, it causes high render times and larger memory footprints. To address these issues, we fuse the RFs into a single representation.

    
    
    }
    
    \subsection{Fusing Radiance Fields}
    {
        \label{sub_sec:fusion}
        Rendering two or more radiance fields simultaneously for composition is computationally expensive and cannot be scaled as the memory and computation increase linearly with the number of radiance fields involved in composition. To this end, we propose a method that quickly distills from multiple RFs to a single representation. The resulting representation is as compact and efficient as a single RF. We leverage the already learned 3D information to take losses in 3D, which leads to faster learning (distillation).

        Let the two radiance fields to be composed be $T_1$ and $T_2$ (teacher1 and teacher2), and the final fused radiance field be $S$ (student). For brevity, let us assume we do not apply any rigid transformation on the radiance fields.
        We shoot a ray through both $T_1$ and $T_2$ and sample points on the ray. The points with low density [\tikzcircle[light_red, fill=light_red]{2pt}] are pruned out while the ones with high density [\tikzcircle[red, fill=red]{2pt}] are utilized. The union of these selected points from $T_1$ and $T_2$ acts as our training set for the student $S$ [\tikzcircle[blue, fill=blue]{2pt}]. We query the three radiance fields $(T_1, T_2, S)$ for their density, alpha and color $(\sigma, \alpha, c)$ at every training sample point location. We apply a \textit{Supervised loss} $||\tikzcircle[red, fill=red]{2pt} - \tikzcircle[blue, fill=blue]{2pt}||_2$ to the color and density values of the student ($S$) against the corresponding teachers ($T_1$, $T_2$) at every training point selected above.
        This supervised distillation will fuse the composition into one single scene.
        
        As a final stage, we render the RGB values for the rays by accumulating the individual color values weighted by the activated volumetric density using Volumetric Rendering Equation\cite{volume_tr} and take an alpha-composited \textit{RGB pixel loss} $||\sqbox{light_red} - \sqbox{light_blue}||_2$ for a few iterations. This helps smooth the result around the boundaries of the inserted object.
        
    }
    \subsection{Fast convergence}
    {   
        \label{sub_sec:fast_conv}
        
        To obtain a single representation of a composed radiance field, one could use the traditional \textit{RGB loss} against the rays from composed RFs or rendered views extracted from the composition. But this would essentially be retraining and would amount to the same time as training an RF from scratch. However, since we have 3D information from already-trained RFs, we can leverage the \textit{supervised losses} employed at every sampled point, which achieves faster convergences. The augmented convergence is due to 3D distillation. Pruning of low-density points suggested in \cref{sub_sec:fusion} further speeds up the process.
        
        Additionally, it is often the case that during the composition of radiance fields, one of the scenes is in the majority (dubbed as a background scene). Initializing the student with the background scene significantly speeds up the distillation process. Hence, we initialize the student representation $S$ with the weights of the background scene (one of the dominant teachers $T_i$). The reduction in time in the case of our distillation-based fusion against total retraining is $3\times$.
    }
    
}

    \begin{figure}[hpt]
    \centering
    \begin{minipage}{0.49\linewidth}
        \subcaptionbox{\label{fig:stump_comp}Naive Composited}
        {
        \begin{tikzpicture}
            \node[anchor=south west,inner sep=0] (image) at (0,0)
            {
            \begin{overpic}[width=\linewidth]{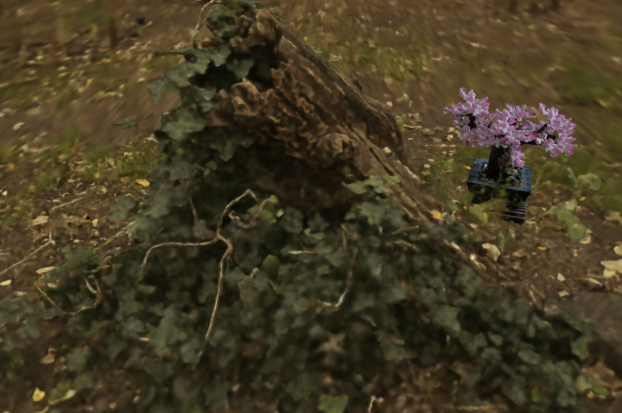}
            \put (1, 1) {\small\color{cyan}\textbf{{Time: 4s, Memory: 50MB}}}
            \end{overpic}
            };
            \begin{scope}[x={(image.south east)},y={(image.north west)}]
                \draw[cyan, thick] (0.7,0.4) rectangle (0.93,0.8);
            \end{scope}
        \end{tikzpicture}
        }
    \end{minipage}
    \begin{minipage}{0.49\linewidth}
        \centering
        \subcaptionbox{\label{fig:stump_fused}Ours FusedRF}
        {
        \begin{tikzpicture}
            \node[anchor=south west,inner sep=0] (image) at (0,0)
            {
            \begin{overpic}[width=\linewidth]{src/images/bonsai_stump_composed.png}
            \put (1,1) {\small\color{cyan}\textbf{{Time: 2s, Memory: 25MB}}}
            \end{overpic}
            };
        \end{tikzpicture}
        }
    \end{minipage}
    \begin{minipage}{0.49\linewidth}
        \centering
        \subcaptionbox{\label{fig:garden_comp}Naive Composited}
        {
        \begin{tikzpicture}
            \node[anchor=south west,inner sep=0] (image) at (0,0)
            {
            \begin{overpic}[width=\linewidth]{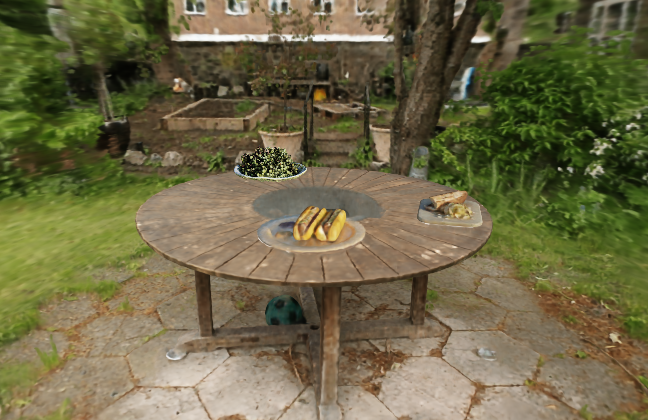}
            \put (1,1) {\small\color{lime}\textbf{{Time: 8s, Memory: 100MB}}}
            \end{overpic}
            };
            \begin{scope}[x={(image.south east)},y={(image.north west)}]
                \draw[lime,thick] (0.4,0.38) rectangle (0.6,0.52);
                \draw[lime,thick] (0.35,0.55) rectangle (0.5,0.68);
                \draw[lime,thick] (0.63,0.43) rectangle (0.77,0.57);
            \end{scope}
        \end{tikzpicture}
        }
    \end{minipage}
    \begin{minipage}{0.49\linewidth}
        \centering
        \subcaptionbox{\label{fig:garden_fused}Ours FusedRF}
        {
        \begin{overpic}[width=\linewidth]{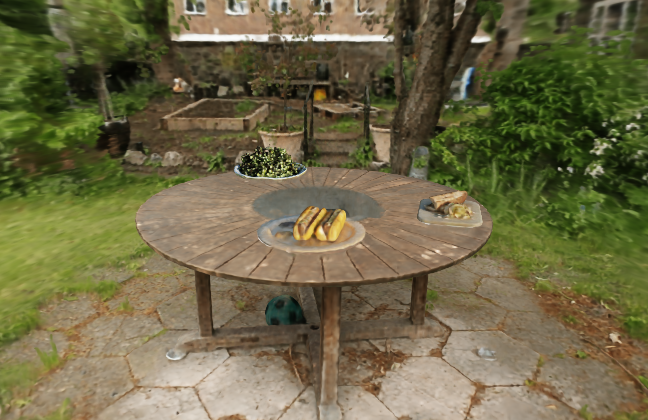}
        \put (1,1) {\small\color{lime}\textbf{{Time: 2s, Memory: 25MB}}}
        \end{overpic}
        }
    \end{minipage}
    \caption{\label{fig:comp_fuse} The figure shows results of compositing multiple RF into a single scene, (left column \cref{fig:garden_comp,fig:stump_comp}) shows results of composition (right column \cref{fig:garden_fused,fig:stump_fused}) shows results of our FusedRF. Respective memory footprints and rendering times are mentioned in the insets. \cref{fig:stump_comp,fig:stump_fused} are two scenes picked from \cite{mipnerf360} and \cref{fig:garden_comp,fig:garden_fused} is combination of \cite{mipnerf360} and synthetics scenes.}
\end{figure}
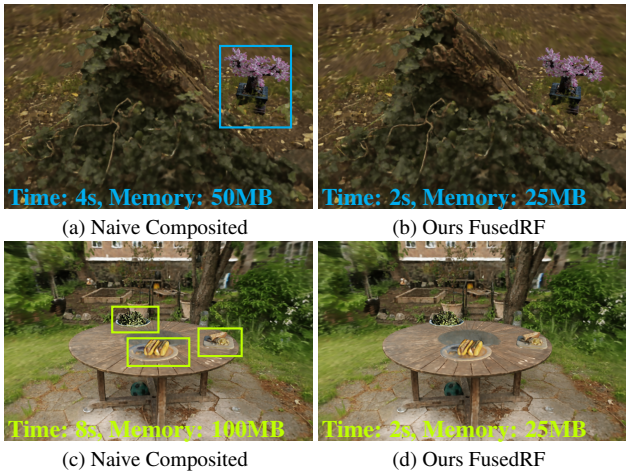
\section{Results}
{
    
    \label{sec:results}
    \paragraph*{Performance:} To validate the performance of our method, we provide Render times and Memory demands against other means of composition, namely 1) Neural-RF, 2) Explicit lattice structures 3) Fused Representation. We tabularize these results in the \cref{tab:memory_timings}. It can be observed that the render times and memory footprints increase linearly in the case of Neural-RFs\cite{nerf} and Explicit lattice representations\cite{plenoxels, tensorf, dvgo, dvgov2} with the increase in the number of scenes used for composition. While it is possible to composite multiple scenes when leveraging Neural-RF, rendering times are a strong limitation, specifically when employed in the case of XR applications. On the other hand, Explicit Lattice representations demand a large amount of memory, leading to infeasibility. On the other hand, our FusedRF representation alleviates the issues by fusing the compositions iteratively, constraining memory, and rendering budgets to that of a single RF.

    \paragraph*{Quantitive Results:} Along with maintaining tighter memory and rendering budgets, our proposed FusedRF representation also retains the quality of composited scenes. To validate this, we provide quantitive metrics of our FusedRF representation against the naive composition(Refer \cref{tab:psnr_table}). This validates the representative capacity of our FusedRF representation.

    \paragraph*{Qualitative Results:} The Qualitative results of our method are presented in the \cref{fig:teaser,fig:comp_fuse}.
    \begin{table}[htp]
  \setlength\tabcolsep{1.0pt} 
  \centering
  \begin{tabular}{lcccc}
    \toprule
    \multirow{2}{*}{\textbf{\#\ }} & \textbf{Neural Based} & \multicolumn{3}{c}{\textbf{Voxel Based}} \\
    \cmidrule(lr){2-2}
    \cmidrule(lr){3-5}
    & \textbf{NeRF} & \textbf{DVGO} & \textbf{TensoRF} & \textbf{Ours}\\
    \midrule
    1 & 10 M / 20s  & 800 M / 2.01s & 25 M  / 2.05s  & 25 M / 2.04s  \\
    2 & 20 M / 40s  & 1.6 G / 4.11s   & 50 M  / 4.25s  & 25 M / 2.05s \\ 
    4 & 40 M / 80s  & 3.2 G / 8.58s   & 100 M  / 8.7s  & 25 M / 2.04s  \\
    8 & 80 M / 160s & {\color{red}OOM}  & 200 M / 17.2s & 25 M / 2.04s  \\ \bottomrule
  \end{tabular}
  \caption{Rendering composition of RFs is {\em slow} and {\em memory intensive} as the number of scenes increases. Our proposed FusedRF performs fusion once, maintaining the memory and computing constant even when the number of composed scenes increases. (M: MB, G: GB, s: seconds). Experimented on RTX 3060 Ti (8GB). \label{tab:memory_timings}}
\end{table}
    \begin{table}[htb]
    \centering
    \begin{tabular}{ccc}
    \toprule
    Scene & Ours (w/o RGB) & Ours (Full) \\
    \midrule
    Figure \ref{fig:teaser}  & 36.71 & 39.89\\
    Figure \ref{fig:stump_fused} & 37.20 & 40.32 \\
    Figure \ref{fig:garden_fused} & 35.18 & 38.78\\
    \bottomrule
    \end{tabular}
    \caption{This table shows the PSNR of the images from some scenes in the paper. Please note that the PSNR reported is of the FusedNeRF images against composed NeRF images.}
    \label{tab:psnr_table}
\end{table}

}

    \section{Conclusion}
{
    \label{sec:conclusion}
    We present FusedRF, a method to create a single RF representation for a scene composed of multiple RFs. This reduces the memory and rendering overheads without degradation of quality. We showed our method over TensoRF \cite{tensorf} representation here. However, our method can be extended to any RF representation that uses explicit 3D lattices like InstantNGP, DVGO, Plenoxels, etc. \cite{instantngp, dvgo, plenoxels}. As our method provides tighter memory and rendering budgets, using our FusedRF in XR applications like \cite{fovnerf, rt_nerf} can facilitate the composition of multiple RFs while maintaining real-time results. The supplementary video provides an overview of our method, the multiview visualization, and the iterative addition of scenes in our results.
}


    {\small
        \bibliographystyle{ieee_fullname}
        \bibliography{main}
    }
}
\end{document}